\documentclass[journal,twoside,web]{ieeecolor}
\usepackage{tmi}
\usepackage{cite}
\usepackage{amsmath,amssymb,amsfonts}
\usepackage{algorithmic}
\usepackage{graphicx}
\usepackage{textcomp}
\usepackage{pdfpages}
\usepackage{multicol}
\usepackage{booktabs}
\usepackage{multirow}
\usepackage{bigstrut}
\usepackage{xspace}
\usepackage{comment}
\usepackage[table,xcdraw]{xcolor}
\usepackage{algorithmic,algorithm}
\RequirePackage{CJKnumb}
\def\BibTeX{{\rm B\kern-.05em{\sc i\kern-.025em b}\kern-.08em
  T\kern-.1667em\lower.7ex\hbox{E}\kern-.125emX}}
\begin{document}
\title{FPL+: Filtered Pseudo Label-based Unsupervised Cross-Modality Adaptation for 3D Medical Image Segmentation}
\author{Jianghao Wu, Dong Guo, Guotai Wang, Qiang Yue, Huijun Yu, Kang Li, Shaoting Zhang
\thanks{This work was supported by the National Natural Science Foundation of China (62271115), National Key Research and Development Program of
China (2020YFB1711500),  Fundamental
Research Funds for the Central Universities (ZYGX2022YGRH019),
 and Sichuan Province International Science, Technology and Innovation Cooperation Foundation (2022YFH0004). (Jianghao Wu and Dong Guo contributed equally to this work. Corresponding authors: Guotai Wang, Shaoting Zhang)} 
\thanks{Jianghao Wu, Dong Guo, Guotai Wang and Huijun Yu are with the School of Mechanical and Electrical Engineering, University of Electronic Science and Technology of China, Chengdu, 611731, China. Jianghao Wu and Guotai Wang are also with Shanghai AI laboratory, Shanghai, 200030, China (e-mail: guotai.wang@uestc.edu.cn).}
\thanks{Qiang Yue is with the Department of Radiology, West China Hospital, Sichuan University, Chengdu, 610041, China}
\thanks{Kang Li is with West China Biomedical Big Data Center, West China Hospital, Sichuan University, Chengdu, 610041, China}
\thanks{Shaoting Zhang is with the School of Mechanical and Electrical Engineering, University of Electronic Science and Technology of China, Chengdu, 611731, China, and also with SenseTime Research, Shanghai, 200233, China (e-mail: zhangshaoting@uestc.edu.cn).}
}
\maketitle

\begin{abstract}
Adapting a medical image segmentation model to a new domain is important for improving its cross-domain transferability, and due to the expensive annotation process, Unsupervised Domain Adaptation (UDA) is appealing where only unlabeled images are needed for the adaptation.
Existing UDA methods are mainly based on image or feature alignment with adversarial training for regularization, and they are limited by insufficient supervision in the target domain. In this paper, we propose an enhanced Filtered Pseudo Label (FPL+)-based UDA method for 3D medical image segmentation. It first uses cross-domain data augmentation to translate labeled images in the source domain to a dual-domain training set consisting of a pseudo source-domain set and a pseudo target-domain set. To leverage the dual-domain augmented images to train a pseudo label generator, domain-specific batch normalization layers are used to deal with the domain shift while learning the domain-invariant structure features, generating high-quality pseudo labels for target-domain images. We then combine labeled source-domain images and target-domain images with pseudo labels to train a final segmentor, where image-level weighting based on uncertainty estimation and pixel-level weighting based on dual-domain consensus are proposed to mitigate the adverse effect of noisy pseudo labels. Experiments on three public multi-modal datasets for Vestibular Schwannoma, brain tumor and whole heart segmentation show that our method surpassed ten state-of-the-art UDA methods, and it even achieved better results than fully supervised learning in the target domain in some cases.
\end{abstract}


\section{Introduction}
\label{sec:introduction}

\IEEEPARstart {D}{eep} learning has revolutionized the field of medical image segmentation, enabling accurate segmentation of various anatomical structures and lesions~\cite{roth2018deep}. For example, algorithms for glioma segmentation have improved dramatically in the last decade, and have achieved results that are close to those of manual segmentation~\cite{baid2021rsna}.
The same phenomenon occurs for Vestibular Schwannoma segmentation, where Convolutional Neural Networks (CNN) have achieved expert-level performance~\cite{wang2019automatic}. However, medical images often have multiple modalities that are different domains with a significant domain gap between them, such as contrast-enhanced T1 (ceT1) and high-resolution T2 (hrT2) Magnetic Resonance (MR) imaging in vestibular schwannoma segmentation. Models trained with one modality often perform poorly on images from another modality. Additionally, manually annotating medical images for each modality is time-consuming and laborious. Therefore, it is impractical to either directly apply a trained model for inference in a new modality, or train a model from labeled images in each modality respectively. To deal with this problem, this work aims to adapt a model trained with one modality to a different modality, so as to improve its performance on the new modality and avoid annotations for the target modality.  

Domain adaptation (DA) is a promising solution to address the problem of dramatic performance degradation across modalities at inference time. It attempts to establish a mapping between the source and target domains so that models trained in the source domain can perform well in the target domain. Early domain adaptation methods necessitate annotations not only in the source domain but also to some extent in the target domain. A naive method is to fine-tune a pre-trained model with annotated images in the target domain~\cite{wang2018interactive}. 
Semi-supervised DA leverages a small set of annotated images and many unannotated ones for adaptation~\cite{Donahue_2013_CVPR, gu2022contrastive}. However, these methods require annotations in the target domain, which is difficult and expensive to obtain for 3D medical images.

Unsupervised Domain Adaptation (UDA) has emerged as a promising technique to address the challenges posed by domain shift in medical image segmentation without relying on labeled target data. Various methods have been proposed to deal with this problem by aligning the source and target domains in terms of image appearance, feature distribution, or output structure. 
Some approaches, exemplified by CycleGAN~\cite{CycleGAN2017} and Contrastive Unpaired Translation (CUT)~\cite{park2020cut}, focus on aligning image appearance between the source and target domains. However, these methods may introduce distortions to the anatomical structure of the images, which can hinder accurate segmentation~\cite{wu2023tiss}.
Tzeng et al.~\cite{tzeng2014deep} and Long et al.~\cite{long2016unsupervised} focused on alignment at the feature level. CycADA~\cite{hoffman2018cycada}, SIFA~\cite{chen2020unsupervised}, and SymD~\cite{han2021deep} aim to align the domains at both the image and feature levels. ADVENT~\cite{vu2019advent} focuses on alignment at the output level that encourages predictions in the target domain to follow the same distribution as labels in the source domain. However, such output alignment may be challenging in cases with significant domain shifts. 
In addition, these methods are mainly proposed for 2D image segmentation, while most medical images are 3D volumes, and dealing with them is more challenging. 

Pseudo labels are widely used for training segmentation models where the annotations are not available or weak, such as in the scenario of semi- and weakly-supervised segmentation~\cite{luo2022scribble,li2020self,lin2022calibrating,pymic2023}. These methods demonstrate that pseudo labels can effectively address the issue of limited annotations by providing more supervisions. However, their application in the context of UDA has been rarely investigated. This is primarily due to the significant domain shift between the source and target domains that makes it challenging to generate reliable pseudo labels. Though pseudo labels can be obtained by models trained in the source domain, they often contain substantial noise, which can mislead the training of a segmentation model in the target domain.

In this work, we propose an enhanced Filtered Pseudo Label (FPL+)-based framework for UDA in 3D medical image segmentation. First, a Cross-Domain Data Augmentation (CDDA) is proposed to augment labeled source-domain images to dual-domain training data with a pseudo source-domain set and a pseudo target-domain set that share the same labels. Then, a Dual-Domain pseudo label Generator (DDG) with dual-domain batch normalization is proposed to learn from the augmented dual-domain images, providing high-quality pseudo labels for the target-domain training set. With the labeled source-domain images and target-domain images with pseudo labels, we further train a final segmentor, where unreliable pseudo labels are suppressed by image-level and pixel-level weighting for robust learning. The proposed CDDA-based pseudo label generator can effectively mitigate the domain gap and obtain accurate pseudo labels in the target domain. By training from both the source-domain and target-domain images, the final segmentor can better learn domain-invariant features that improve performance in the target domain. The contributions of this work are summarized in three aspects:

\begin{itemize}
	\item We propose a novel UDA framework named FPL+ for cross-modality 3D medical image segmentation based on generating high-quality pseudo labels in the target domain and noise-robust learning, which is different from existing methods using image, feature or output alignment that are often proposed for UDA in 2D segmentation. 
	\item We introduce a novel pseudo label generation method based on Cross-Domain Data Augmentation (CDDA) and Dual-Domain pseudo label Generator (DDG), where CDDA augments the labeled source-domain images to a pseudo source-domain set and a pseudo target-domain set, and the DDG is based on dual batch normalization to learn from the augmented dual-domain training set, which effectively mitigates the large cross-modality domain gap.
	\item We propose a joint learning method to train a final segmentor from a combination of the labeled source-domain images and target-domain images with pseudo labels, where image-level weighting based on uncertainty estimation and pixel-level weighting based on dual-domain consensus are introduced for noise-robust learning. 
\end{itemize}

This work is a substantial extension of our preliminary conference publication~\cite{wu2022fpl}. In the preliminary study, we used Generative Adversarial Networks (GAN)-based data augmentation to obtain more pseudo source-domain images to train a pseudo label generator, and image-level weighting is used for learning from the pseudo labels of target-domain images.  The main differences of this work from~\cite{wu2022fpl} include: 1) Instead of augmenting source-domain images only to pseudo source-domain images, the CDDA in this work translates the labeled source domain data into dual-domain training data consisting of a pseudo source-domain set and a pseudo target-domain set; 2) The pseudo label generator learns only from source-domain and pseudo source-domain images in~\cite{wu2022fpl}, while a DDG is introduced in this work to learn from the dual-domain augmented training set; 3) The final segmentor in~\cite{wu2022fpl} is trained with target-domain images with pseudo labels only, while this work proposes joint training that additionally leverages labeled source-domain images to train the final segmentor in the target domain; 4)~In addition to image-level weighting, pixel-level weighting is further introduced to learn from reliable pseudo labels;
5)~Compared with mono-directional cross-modality adaptation for Vestibular Schwannoma segmentation in~\cite{wu2022fpl}, the method in this work is further validated with a glioma segmentation dataset, and bidirectional cross-modality adaptation is implemented in the experiment.  The results showed that our method outperforms ten state-of-the-art (SOTA) UDA methods. The code is available online\footnote{https://github.com/HiLab-git/FPL-plus}. 

\begin{figure*}
    \centering
    \centerline{\includegraphics[width=16cm]{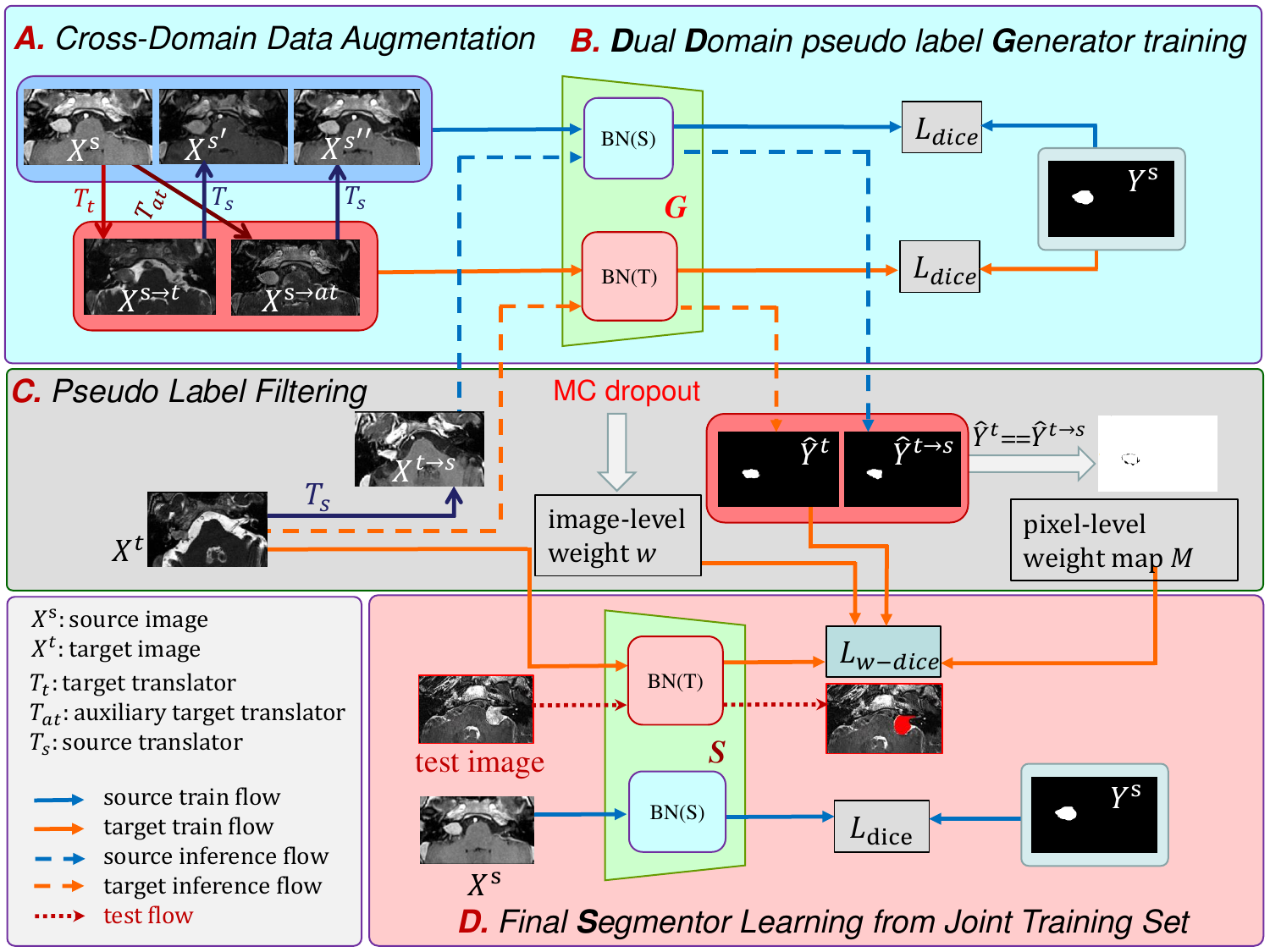}}
    \caption{Overview  of  our enhanced Filtered Pseudo Label (FPL+)-based framework for cross-modality UDA. A) Cross-Domain Data Augmentation (CDDA) augments the labeled source-domain images into a dual-domain training set. 
    B) A Dual-Domain pseudo label Generator (DDG) is trained with the augmented dual-domain training images using dual batch normalization layers. The final segmentor is jointly trained with source-domain and target-domain images (D), where pseudo labels for the target domain are filtered based on image-level and pixel-level weighing (C). For testing, the target-domain image is inferred directly using the trained final segmentor with target-domain batch normalization. }
    \label{fig:method}
\end{figure*}

\section{Related Work}
\subsection{Unsupervised Domain Adaptation}
Unsupervised Domain Adaptation (UDA) aims to transfer knowledge from labeled data in a source domain to an unlabeled target domain. It typically operates by aligning the source and target domains at different levels.
Firstly, image appearance alignment methods such as CycleGAN~\cite{CycleGAN2017} and CUT~\cite{park2020cut} learn a mapping between the source and target domains for cross-domain image translation while preserving the content of images as much as possible. Secondly, feature alignment methods aim to minimize the distance of feature distribution between the source and target domains to learn domain-invariant representations~\cite{dou2019pnp,wu2020cf}. For example, Dou et al.~\cite{dou2019pnp} proposed to implicitly align the feature spaces of source and target domains at multiple scales with an adversarial loss, and Wu et al.~\cite{wu2020cf} proposed a characteristic function distance to explicitly reduce the distribution discrepancy between the two domains. 
Thirdly, output alignment methods align the shape or structure of the predictions between the source and target domains~\cite{vu2019advent}. This is particularly relevant in medical image segmentation, where the shape and structure of the target organ or lesion can vary between different imaging modalities or datasets~\cite{wu2022fpl,yao2022unsupervised}. However, most of these methods are primarily designed for segmenting 2D medical images, which have limited performance on 3D medical image segmentation. DAR-Net~\cite{yao2022novel} combines a 2D style transfer network and a 3D segmentation network to deal with 3D medical images. However, it can hardly obtain realistic style transfer due to lesions or limited training images, and there is still a domain gap between synthetic and real target 3D images, leading to a limited performance. 

\subsection{Learning from Noisy Labels}
Pseudo label learning has been widely used in medical image segmentation to deal with unannotated images or pixels in the training set~\cite{li2020self,lin2022calibrating,wu2022mutual,luo2022scribble,pymic2023}. However, as pseudo labels are obtained from an insufficiently trained model, they are often noisy due to incorrect predictions and may limit the model's performance. 
Noisy labels may also be from imperfect manual annotations, so noise-robust learning methods have been proposed for dealing with both noisy pseudo labels and imperfect manual annotations.
Various techniques have been proposed to tackle this issue, including noise robust loss functions~\cite{ghosh2017robust,zhang2018generalized,wang2020noise}, label correction methods~\cite{xu2022anti,Wang2022semi}, and training multiple networks~\cite{han2018co,zhang2020robust,yang2022learning}. 
For noise-robust loss functions, Zhang et al.~\cite{zhang2018generalized} proposed a generalized cross entropy loss, and Wang et al.~\cite{wang2020noise} proposed a noise-robust Dice loss used in an adaptive mean teacher framework.  
Label correction aims to refine noisy labels during training. Xu et al.~\cite{xu2022anti} proposed mean-teacher-assisted confident learning to select and refine low-quality pseudo labels. Wang et al.~\cite{Wang2022semi} proposed iterative refinement of pseudo labels based on uncertainty-guided conditional random fields.
For training multiple networks, Co-teaching~\cite{han2018co} used two neural networks to learn from each other, which reduces the risk of over-fitting on noise by a single network. Yang et al.~\cite{yang2022learning} proposed a dual-branch network to distinguish high-quality and low-quality pseudo labels and leverage them with different strategies for COVID-19 pneumonia lesion segmentation.
Zhang et al.~\cite{zhang2020robust} proposed a tri-network learning framework, where each two networks select high-quality pseudo labels to supervise the other. However, these methods are computationally expensive for 3D segmentation, and it is still challenging to learn from noisy pseudo labels for UDA due to their low quality.

\section{Method}

Our proposed FPL+ framework is illustrated in Fig.~\ref{fig:method}. To achieve cross-modality UDA for 3D medical image segmentation, it first obtains high-quality pseudo labels for training images in the target domain, and then trains a segmentation model in that domain by learning from pseudo labels. To improve the performance of the pseudo label generator, we first propose Cross-Domain Data Augmentation (CDDA) that augments labeled source-domain images into a dual-domain dataset consisting of a  pseudo source-domain set and a pseudo target-domain set with the same set of labels. Then, a Dual-Domain pseudo label Generator (DDG) learns from the dual-domain augmented images to produce high-quality pseudo labels for training images in the target domain. To train a final segmentor, we introduce joint training from the labeled source-domain images and target-domain images with pseudo labels, and propose image-level weighting based on size-aware uncertainty estimation and pixel-level weighting based on dual-domain consensus to mitigate the adverse effects of unreliable pseudo labels.

\subsection{Cross-Domain Data Augmentation}
Let $\mathcal{D}_s$ and $\mathcal{D}_t$ denote a set of labeled source-domain images and a set of unlabeled target-domain images, respectively. Let $\it{X^s_i}$ and $\it{X^t_j}$ denote the $i$-th image from $\mathcal{D}_s$ and the $j$-th image from $\mathcal{D}_t$, respectively, where the label of $\it{X^s_i}$ is $\it{Y^s_i}$. Note that the source domain and target domain are from different patient groups, i.e., $\it{X^s_i}$ and $\it{X^t_j}$ are unpaired.
Due to the domain shift between $\mathcal{D}_s$  and $\mathcal{D}_t$, training a model with $\mathcal{D}_s$ to generate pseudo labels for $\mathcal{D}_t$ will lead to a poor performance. In order to improve the quality of pseudo labels for $\mathcal{D}_t$, we propose Cross-Domain Data Augmentation (CDDA) to augment $\mathcal{D}_s$ before training the pseudo label generator.

\begin{figure}
    \centering
    \centerline{\includegraphics[width=9cm]{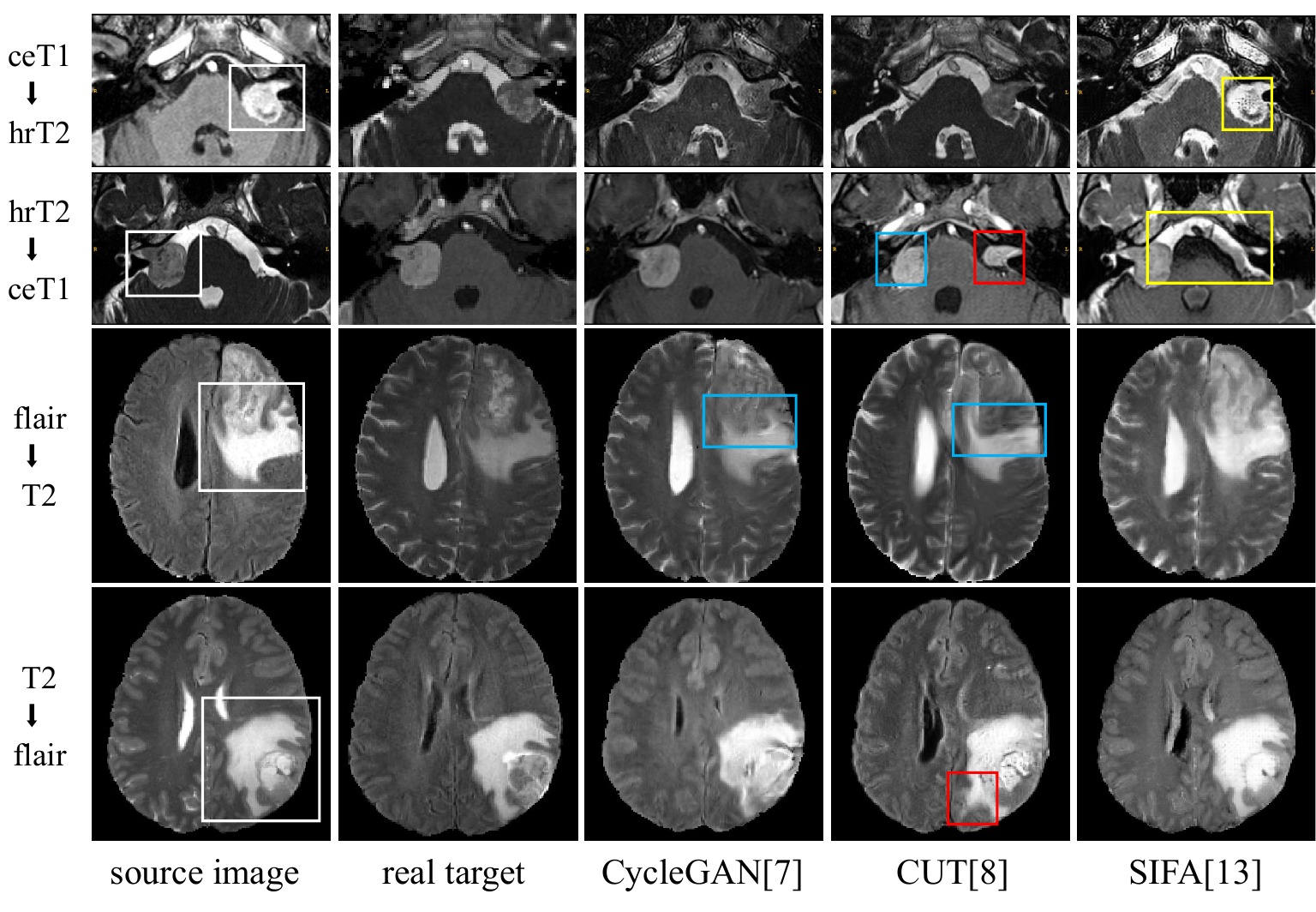}}
    \caption{Visualization of different methods for image translation trained with unpaired source and target domain images. White boxes represent the tumor region for segmentation. Blue boxes show synthetic regions with shrunk tumors. Red boxes represent artifact tumor regions. Yellow boxes represent insufficient style translation. }
    \label{fig:generation}
\end{figure}

Specifically, we utilize an image style translator $T_t$ to translate a labeled source-domain image $X^s_i$ into a pseudo target-domain image $X_i^{s \rightarrow t}$ = $T_t(X^s_i)$, and use another image style translator  $T_s$ to translate $X_i^{s \rightarrow t}$ back to the source domain, leading to a pseudo source-domain image $X_i^{s'}$ = $T_s(X_i^{s \rightarrow t})$. Note that $T_t$ and $T_s$ are often trained jointly for learning from unpaired training sets, as used in CycleGAN~\cite{CycleGAN2017}. As the training sets are unpaired, it is difficult to make $X_i^{s \rightarrow t}$ and $X^{s'}_t$ exactly match the ground truth target-modality and source-modality images, respectively. As a result, the images may have some structure distortions after style translation. Fig.~\ref{fig:generation} presents some examples of translated images obtained by different methods, including CycleGAN~\cite{CycleGAN2017}, CUT~\cite{park2020cut} and SIFA~\cite{chen2020unsupervised}. It shows that the translated images may have different quality issues, such as shrunk and artefact tumors, or insufficient style translation. 

To enhance the diversity of the training images and reduce the risk of over-fitting to the structure distortions obtained by the image translator $T_t$ when training the pseudo label generator, we introduce an auxiliary target style translator $T_{at}$ that shares the same architecture as $T_t$, and its weights are obtained from a different checkpoint during the training process of $T_t$, as we observed that the translator at different checkpoints can lead to some different local details. Unlike training CUT~\cite{park2020cut} as a second translator in FPL~\cite{wu2022fpl}, $T_{at}$ does not need an extra training process, and can provide data augmentation by translating a source domain image $X^s_i$ into a different pseudo target-domain image $X_i^{s \rightarrow at}$. We also translate $X_i^{s \rightarrow at}$ back to a pseudo source-domain image $X_i^{s''}$=  $T_s(X_i^{s \rightarrow at})$. 
As a result, for each labeled source-domain image $X^s_i$, we obtain four augmented images, i.e, two pseudo source-domain images $X^{s'}_i$ and $X^{s''}_i$ and two pseudo target-domain images $X_i^{s \rightarrow t}$ and $X_i^{s \rightarrow at}$, and they share the same segmentation label $Y^s_i$. We denote the augmented source-domain training set as $\mathcal{D}_{ss} = \{ X_i^s, X_i^{s'}, X_i^{s''}\}$, and the pseudo target-domain training set as $\mathcal{D}_{st} = \{ X_i^{s \rightarrow t}, X_i^{s \rightarrow at } \}$, respectively. Then $\mathcal{D}_{ss}$ and $\mathcal{D}_{st}$ are used together to train the pseudo label generator. 

In this work, the image translators $T_s$ and $T_t$ are implemented based on CycleGAN~\cite{CycleGAN2017} with two discriminators $D_s$ and $D_t$ for the two domains, respectively. The training involves two adversarial losses $\mathcal{L}_{gan}^t$, $\mathcal{L}_{gan}^s$ and a cycle consistency loss $\mathcal{L}_{cyc}$. The target-domain adversarial loss $\mathcal{L}_{gan}^t$ is:
\begin{equation}
\label{equ:gan_t}
\begin{aligned}
    \mathcal{L}_{gan}^t(T_t, D_t) = & \mathbb{ E}_{X^t_{j} \sim \mathcal{D}_t}[\text{log} D_t(X^t_{j})]\\
    + & \mathbb{ E}_{X^s_i \sim \mathcal{D}_s}[\text{log} (1 - D_t(X^{s \rightarrow t}_i))]\\
\end{aligned}
\end{equation}

The source-domain adversarial loss $\mathcal{L}_{gan}^s$ is defined similarly based on $T_s$, $D_s$ and $X^{s'}_i$, and the consistency loss is:
\begin{equation}
\label{equ:cyc}
\begin{aligned}
    \mathcal{L}_{cyc}(T_{s}, T_{t}) = & 
    \mathbb{E}_{X^s_i\sim \mathcal{D}_s}[{\Vert T_s(T_t(X^s_i)) - X^s_i \Vert}_1] \\ + & \mathbb{E}_{X^t_{j}\sim \mathcal{D}_t}[{\Vert T_t(T_s(X^t_{j})) - X^t_{j} \Vert}_1]
\end{aligned}
\end{equation}

\subsection{Dual-domain Pseudo Label Generator}
After CDDA, the dual-domain augmented training set has more training samples with different appearances and shared segmentation labels for training the pseudo label generator. 
However, images in $\mathcal{D}_{ss}$ and $\mathcal{D}_{st}$ exhibit different modalities, leading to different statistics that make it difficult to use them jointly to train a segmentation model in a standard fully supervised setting~\cite{Gu2022cssda,zhou2022dn}. 

To effectively leverage the augmented training set and deal with the different statistics, we propose a Dual-Domain pseudo label Generator (DDG) $G$ that uses dual batch normalization (Dual-BN)~\cite{chang2019domain} to normalize the features of the source and target domains respectively. Specifically, the features in a certain layer extracted from the source domain are normalized by a source-domain BN layer, and those from the target domain are normalized by a target-domain BN layer~\cite{zhou2022dn}:

\begin{equation}
    \text{Dual-BN}(z_d;d) = \gamma_d \frac{z_d - \mu_d}{\sqrt{\sigma_d^2 + \epsilon}} + \beta_d
\end{equation}
where $z_d$ represents the  features obtained from domain $d$, and
$d \in \{s, t \}$ represents the domain label, i.e., $d=s$ when the input is from $\mathcal{D}_{ss}$ and $d=t$ otherwise.  $\gamma_d$ and $\beta_d$ are learnable parameters, and $\mu_d$ and $\sigma_d$ are the mean and standard deviation of the corresponding domain, respectively. The small constant $\epsilon > 0$ is added to ensure numerical stability.

During training, the BN layers estimate the means and variances of features using exponential moving average with a factor of $\alpha$~\cite{ioffe2015batch}. For Dual-BN, they are given by:
\begin{equation}
    \overline{\mu}_d^{k+1} = (1 - \alpha)\overline{\mu}_d^{k} + \alpha \mu_{d}^{k}
\end{equation}
\begin{equation}
    (\overline{\sigma}_d^{k+1})^2 = (1 - \alpha) (\overline{\sigma}_d^{k})^2 + \alpha (\sigma_{d}^k)^2
\end{equation}
where $\overline{\mu}_{d}^{k}$ and $(\overline{\sigma}_{d}^{k})^2$ are the estimated mean and variance of domain $d$ at iteration $k$, and $\alpha$ is the momentum parameter for the moving average. 

Moreover, the other parameters in $G$ are shared between $\mathcal{D}_{ss}$ and $\mathcal{D}_{st}$, which facilitates the learning of more domain-invariant features by leveraging the dual-domain augmented training set. 
The parameters of $G$ are denoted as ${\theta}_{G} = [\theta, \gamma_{s}, \beta_{s}, \gamma_{t}, \beta_{t}]$, where $\theta$ represents the shared parameters except for those in batch normalization layers. $\gamma_{s}$ and $\beta_{s}$ are the source domain-specific BN parameters, and $ \gamma_{t}$ and $ \beta_{t}$ are  target domain-specific BN parameters, respectively.

Let $M$ and $N$ denote the number of samples in $\mathcal{D}_{ss}$ and $\mathcal{D}_{st}$, respectively. 
For a sample $X^s_m \in \mathcal{D}_{ss}$, the prediction is denoted as
$\Tilde{Y}_m^s = G(X^s_m; \theta, \gamma_{s}, \beta_{s})$, and for a sample $X^t_n  \in \mathcal{D}_{st}$, the prediction is $\Tilde{Y}_n^t = G(X_n^t; \theta, \gamma_{t}, \beta_{t})$. The ground truth of $X^s_m$ and $X^t_n$ are denoted as $Y^s_m$ and $Y^t_n$, respectively. The loss function to train $G$ on $\mathcal{D}_{ss}$ and $\mathcal{D}_{st}$ is:
\begin{equation}
    \begin{aligned}
       \mathcal{L}(\theta_G) = \frac{1}{M}\sum_{m = 1}^M \mathcal{L}_{dice}(Y^s_m, \tilde{Y}_m^s) + \frac{1}{N}\sum_{n = 1}^N \mathcal{L}_{dice}(Y^t_n, \Tilde{Y}_n^t)
    \end{aligned}
\end{equation}
where $\mathcal{L}_{dice}$ is the Dice loss  for supervised segmentation.

\subsection{Pseudo Label Filtering}
After training the pseudo label generator $G$, a pseudo label  for an image $X^t_{j} \in \mathcal{D}_t$ in the target domain is obtained by $P^t_{j} = G(X^t_{j}; \theta, \gamma_{t}, \beta_{t})$, where the BN layers use the target domain-specific parameters. As the quality of these pseudo labels varies in different samples and image regions, directly using all pseudo labels as ground truth for training may mislead the final segmentor. Therefore, we propose pseudo label filtering based on image-level and pixel-level weighting for robust learning.

\subsubsection{Image-level Weighting based on Size-aware Uncertainty Estimation}
For image-level weighting, 
uncertainty estimation based on Monte Carlo (MC) dropout~\cite{Gal2016}  is a widely used method, and pseudo labels with a larger uncertain region are likely to be unreliable. However, in segmentation tasks, uncertain regions are often located at edges of the targets, resulting in higher overall uncertainty for cases with larger targets, which may neglect images with small targets, especially for tumor segmentation with various sizes. 
To deal with this problem, we propose a size-aware uncertainty estimation method for reliable image-level weighting of pseudo labels. 

Firstly, when obtaining pseudo labels for the target-domain training set, we enable dropout layers of $G$ and make $K$ consecutive predictions for each case, leading to $K$ different probability maps for the same case. 
Let $\bar P_{j}^t$ denote the average result across these probability maps for image $X_{j}^t$, and the pseudo label $\hat{Y}^t_{j}$ is obtained by taking an argmax on $\bar P^t_{j}$. For each pixel, we calculate the variance of the foreground probability across the $K$ predictions, leading to a variance map $V_{j}$ with the same shape as $X^{t}_{j}$. The value of each pixel in $V_{j}$ is summed to get a naive image-level uncertainty $v_{j}$:
\begin{equation}\label{v}
{v_{j}} = \sum_{o} V_{j,o}
\end{equation}
where $V_{j,o}$ is the variance of pixel $o$ in $V_{j}$. 

Secondly, as  $v_{j}$ tends to be biased towards images with large targets, we normalize $v_{j}$ by the estimated size of uncertain region that is denoted as ${\eta}_{j}$. Specifically, we calculate the entropy $E_{j, o}$ of pixel $o$ based on ${\bar P}^t_{j}$. $\eta_{j}$ is defined as:
\begin{equation}\label{eta}
\eta_{j} = \sum_{o} \mathcal{H}(E_{j,o} - e)
\end{equation}
where $e$ is a threshold for pixel-level entropy. $\mathcal{H}(\cdot)$ is the unit step function that takes 0.0 for negative inputs and 1.0 for positive inputs. Then, our proposed image-level uncertainty for the pseudo label $\hat{Y}^t_{j}$ of image $X^t_{j}$ is:

\begin{equation}
u_{j} = \begin{cases}
\frac{v_{j}}{\eta_{j}}, & \text{if } \eta_{j} > 0 \\
u^{*}, & \text{else}  \\
\end{cases}
\end{equation}
where $u^{*}$ represents the maximum value of $u_{j}$ when $\eta_{j} > 0$.

Finally, the image-level weight $w_{{j}}$ for $\hat{Y}^t_{{j}}$ is defined as:
\begin{equation}
    w_{{j}} = \frac{u^{*} - u_{{j}} }{u^{*} - u_{min}},  
\end{equation}
where $u_{min}$ is the minimal value of $u_{{j}}$, and a smaller image-level uncertainty leads to a higher image-level weight. 

\subsubsection{Pixel-level Weighting based on Dual-Domain Consensus}
To further reduce the affect of unreliable predictions at the pixel level, we introduce a dual-domain consensus-based weight map obtained by applying $G$ to 
$X^t_{{j}}$ and its style-translated version.
Specifically, we use $T_s$ to translate $X^t_{{i}}$ in $\mathcal{D}_{t}$ into a pseudo source-domain image $X_{{j}}^{t \rightarrow s} $, and obtain another pseudo label: $\hat{Y}_{{j}}^{t \rightarrow s}=G(X_{{j}}^{t \rightarrow s}; \theta, \gamma_{s}, \beta_{s})$, where the BN layers in $G$ use source-domain-specifc parameters.  We then treat the consensus and discrepancy regions between $\hat{Y}_{{j}}^{t \rightarrow s}$ and $\hat{Y}_{{j}}^{t}$  as reliable and unreliable  predictions, respectively. A weight map $M_{{j}}$ is defined by:

\begin{equation}
    M_{{j}} = [\hat{Y}^t_{{j}} == \hat{Y}^{t \rightarrow s}_{{j}}]
\end{equation}
where we set the pixel-level weight as 1.0 and 0.0 for the consensus and discrepancy, respectively.

After obtaining $w_{{j}}$ and $M_{{j}}$ for $X_{{j}}^t$, we combine them into a single weight map $A_{{j}} = M_{{j}}\cdot w_{{j}}$, which integrates both the image-level and pixel-level weighting in a unified formulation. 
To make the model learn more from reliable information, and to reduce overfitting to unreliable information, a weighted Dice loss $\mathcal{L}_{w-dice}$ is proposed to learn from $\hat{Y}^t_{{j}}$ with $A_{{j}}$:

\begin{equation}\label{eq:w-dice}
    \mathcal{L}_{w-dice}
    = 1-\frac{1}{\mathcal{N}}\frac{ \sum_{n=1}^{\mathcal{N}} 2A_{{{j}},n}\tilde{Y}^t_{{{j}},n}\hat{Y}^t_{{{j}},n}}{ \sum_{n=1}^{\mathcal{N}} A_{{{j}},n}(\tilde{Y}^t_{{{j}},n} + \hat{Y}^t_{{{j}},n}) +\epsilon}
\end{equation}
where $\mathcal{N}$ is the pixel number in $X^t_{{j}}$. $\tilde{Y}^t_{{j}}$ is the prediction for $X^t_{{j}}$. Note that Eq.~\ref{eq:w-dice}
is defined for a binary segmentation task, and it can be easily extended for multi-class segmentation. The image-level weighting and pixel-level weighting are generated only once before the training of the final segmentor. Subsequently, the pseudo labels and the weighting values remain fixed throughout the training iterations. 



\subsection{Final Segmentor Learning from Joint Training Set}
Though the target-domain images $\mathcal{D}_t$ with pseudo labels can be used to train a final segmentor in the target domain, we have labeled source-domain images $\mathcal{D}_s$ at hand, and combining $\mathcal{D}_s$ and $\mathcal{D}_t$ to train the final segmentor can better leverage the knowledge in the source domain to improve its performance.  
Therefore, we propose a dual-domain segmentor $S$ to jointly learn from labeled images in $\mathcal{D}_s$ and images with pseudo labels in $\mathcal{D}_t$.  To deal with the domain shift between $\mathcal{D}_s$ and $\mathcal{D}_t$ for joint learning, $S$ is designed with the same architecture as the pseudo label generator $G$ based on dual-BN layers. Similarly to $G$, the parameters of $S$ are denoted as ${\theta}_{S} = [\theta, \gamma_{s}, \beta_{s}, \gamma_{t}, \beta_{t}]$. The training loss for $S$ is:

\begin{equation}
    \begin{aligned}
       \mathcal{L}_{\theta_S} &= \frac{1}{|\mathcal{D}_s|}\sum_{X^s_i \in  \mathcal{D}_s}  \mathcal{L}_{dice}(Y^s_i, \tilde{Y}^s_i) \\
       &+\frac{1}{|\mathcal{D}_t|}\sum_{X^t_{\text{j}} \in \mathcal{D}_t} \mathcal{L}_{w-dice}(\hat{Y}^t_{{j}}, \tilde{Y}^t_{{j}}, A_{{j}})
    \end{aligned}
\end{equation}
where $\tilde{Y}^s_i=S(X^s_i; \theta, \gamma_{s}, \beta_{s})$ and $\tilde{Y}^t_{{j}}=S(X^t_{{j}}; \theta, \gamma_{t}, \beta_{t})$. $\mathcal{L}_{w-dice}$ is the weighted Dice loss  defined in Eq.~\ref{eq:w-dice}.
To accelerate the training of $S$, we initialize it with the weights of $G$ due to their shared architecture.
During the testing stage in the target domain, as shown in Fig.~\ref{fig:method} (D), we directly use $S$ with the target-domain-specific BN layers for inference.

\section{Experiment}
\label{sec:exp}

\subsection{Datasets and Implementation}
\subsubsection{Vestibular Schwannoma Segmentation Dataset}
We first validated our method on the publicly available Vestibular Schwannoma (VS) segmentation dataset~\cite{shapey2021segmentation}, which includes 3D MRI images from 242 patients. Each patient was scanned by contrast-enhanced T1-weighted (ceT1) and high-resolution T2-weighted (hrT2) MRI, with in-plane resolution around 0.4 mm × 0.4 mm, in-plane size of 512 × 512, and slice thickness of 1.5 mm. We used the two modalities for bidirectional adaptation, i.e., using ceT1 and hrT2 as source and target domains, respectively, and vice versa.  
We randomly split the dataset into 200 patients for training, 14 patients for validation and 28 patients for testing. In the training set, images in one modality of 100 patients were used as the source domain, and images in the other modality of the other 100 patients were used as the target domain. 
We followed the setting of the Cross-modality Domain Adaptation Challenge 2021 (CrossMoDA 2021)~\cite{dorent2023crossmoda} to use validation set in the target domain to tune hyper-parameters, and the testing set was only used in the final inference. For preprocessing, each image was cropped by a cubic box determined by the largest possible range of VS in the training set and expanded by a margin, and normalized by intensity mean and standard deviation.

\begin{table}
  \centering
  \caption{Quantitative comparison of different UDA methods for bidirectional adaptation on vestibular schwannoma segmentation. $\dag$ indicates a significant improvement ($p$-value < 0.05) from the best values obtained by existing methods. }
  \resizebox{9cm}{!}{
    \begin{tabular}{c|c|c|c|c}
    \hline
    \multirow{2}[4]{*}{Method} & \multicolumn{2}{c|}{ceT1 to hrT2} & \multicolumn{2}{c}{hrT2 to ceT1} \bigstrut\\
\cline{2-5}          & Dice (\%) & ASSD (mm) & Dice (\%) & ASSD (mm) \bigstrut\\
    \hline
    w/o DA & 0.00±0.00 & 48.30±5.29 
    & 2.65±8.18 & 31.01±16.61 \bigstrut[t]\\
    labeled target & 88.17±7.81 & 1.03±2.67 & 90.72±12.47 & 0.30±0.53 \\
    {strong upbound} & {89.40±5.89} & {0.64±1.44} & {94.23±2.97} & {0.17±0.16} \bigstrut[b]\\
    \hline
    SIFA~\cite{chen2020unsupervised}  & 69.75±21.54 & 6.01±5.88 & 67.48±20.32 & 6.51±8.89 \bigstrut[t]\\
    AccSeg~\cite{zhou2021anatomy} & 30.95±31.81 & 15.44±10.63 & 37.01±31.97 & 17.06±21.11 \\
    ADVENT~\cite{vu2019advent} & 5.36±9.61 & 35.68±11.49 & 21.94±23.07 & 34.11±15.24 \\
{HRDA}~\cite{hoyer2022hrda} & {6.15±13.38} & {21.69±16.67} &  {17.72±19.74} & {14.69±11.48} \\
{CDAC}~\cite{wang2023cdac} & {0.32±1.38} & {25.39±11.00} &{2.98±8.13} & {35.54±18.57} \\
{MIC}~\cite{hoyer2023mic} & {54.82±24.55} & {11.84±11.66} & {13.44±22.95} & {30.13±22.37} \\

    CycleGAN~\cite{CycleGAN2017} & 74.36±24.84 & 2.19±4.26 & 65.79±37.20 & 7.09±15.14 \\
    CUT~\cite{park2020cut}   & 73.64±15.57 & 3.96±6.86 & 56.27±31.37 & 9.25±17.14 \\
    DAR-NET~\cite{yao2022novel} & 76.52±21.34 & 3.26±3.69 & 84.29±14.39 & 1.57±2.94 \\
    FPL~\cite{wu2022fpl} & 78.78±17.88 & 1.56±3.91 & 84.90±14.29 & 0.97±1.56 \\
    FPL+ (Ours) & \textbf{82.92±12.22\dag} & \textbf{0.93±1.82} & \textbf{91.98±6.03\dag} & \textbf{0.23±0.26\dag} \bigstrut[b]\\
    \hline
    \end{tabular}%
    }
  \label{tab:sota_vs}%
\end{table}%

\begin{figure*}
    \centering
    \centerline{\includegraphics[width=17cm]{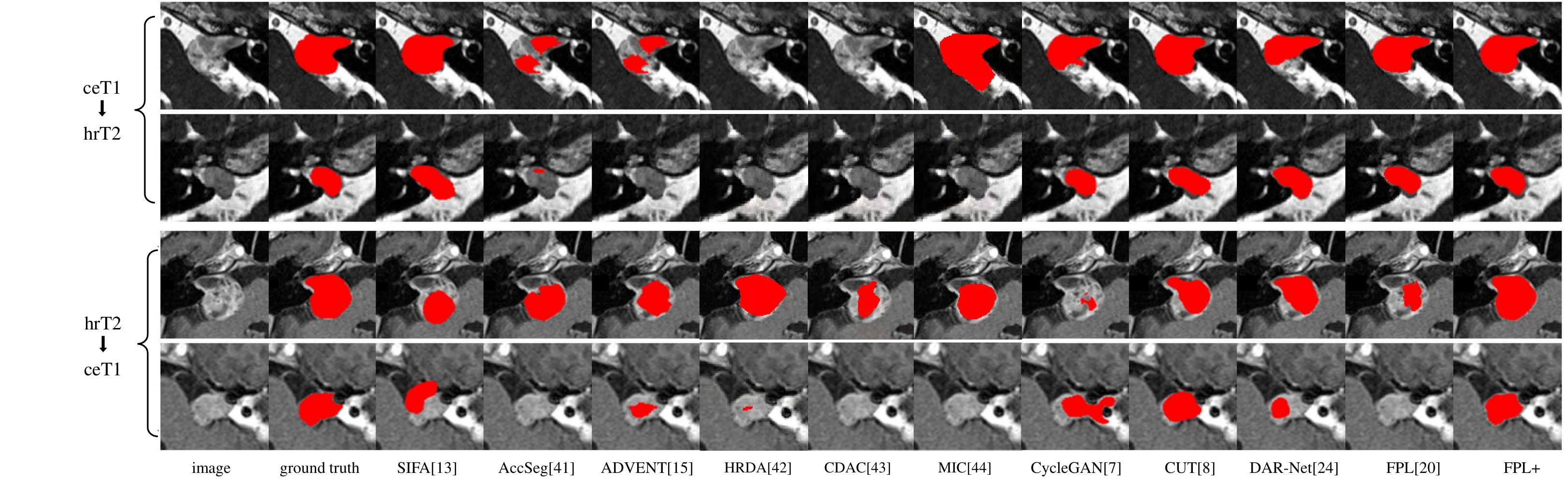}}
    \caption{Visualization of segmentation results obtained by different UDA methods on the VS dataset. 
    }
    \label{fig:seg_vs}
\end{figure*}
\subsubsection{BraTS Dataset}
Our method was also validated on the multi-modal Brain Tumor Segmentation (BraTS) challenge 2020 dataset~\cite{menze2014multimodal}. As the ground truth of the official validation and testing sets are not publicly available, we used the official training set for experiments, which includes spatially aligned MRI scans of four modalities (T1, ceT1, T2, and FLAIR) from 369 patients with a resolution of 1.0 mm$^3$ and an in-plane size of 240 × 240. 
We used T2 and FLAIR images for bidirectional adaptation, and aimed to segment the whole tumor. In each direction, we used images in one modality from 143 patients as the source domain, and images in the other modality from another 143 patients as the target domain. 42 (21 for each direction) and 41 images in the target domain were used for validation and testing, respectively. 
For preprocessing, the intensity of each modality was normalized by the mean and standard deviation. We removed the first and last 20 slices of each volume along the z-axis as they do not contain tumors.

\subsubsection{MMWHS Dataset}
{The MMWHS dataset (Multi-Modality Whole Heart Segmentation Challenge 2017)~\cite{luo2022mathcal} consists of 20 3D CT scans and 20 3D MRI scans. The segmentation targets include the Ascending Aorta (AA), Left Atrium Blood Cavity (LAC), Left Ventricle Blood Cavity} (LVC), and Myocardium of the Left Ventricle (MYO). Following the experimental setting in in~\cite{xian2023unsupervised}, we designated MRI as the source domain and CT as the target domain. Each domain comprised 14, 2, and 4 volumes for training, validation and testing, respectively. For preprocessing, each volume was cropped by a cubic box defined by the maximum extent of the entire heart, and the intensity values were normalized using the mean and standard deviation.


\subsubsection{Implementation Details}
The pseudo label generator $G$ and final segmentor $S$ were implemented by a modified version of an existing 2.5D network~\cite{wang2019automatic} designed for brain tumor segmentation. It has a U-Net-like structure, where the first two resolution levels used 2D convolutions and the other resolution levels used 3D convolutions. 
We added an extra BN layer to all blocks for dual-domain batch normalization. Both $G$ and $S$ were trained using the Adam optimizer with momentum of 0.9 and an initial learning rate of 10$^{-3}$.  $G$ was trained for 200 epochs, while $S$ was initialized by $G$ and trained for another 100 epochs. The patch sizes were 32$\times$128$\times$128 and 32$\times$192$\times$192 for VS and BraTS, respectively, and the batch size was 4 for VS and 2 for BraTS, respectively. We followed the implementation of CycleGAN to train $T_s$ and $T_t$ using 2D slices.  
The training was conducted for 300 epochs, and we selected the checkpoint at 200th epoch as the weight of $T_{at}$ and the 300th epoch as the weight of the $T_s$ and $T_t$. The hyper-parameter $K$ for Monte Carlo dropout was 5, and $e$ was set to 0.2 in the experiments. \textcolor{black}{For fairness and reproducibility, all our hyper-parameters followed the aforementioned settings, and they were not updated during training.}
We implemented all the experiments using PyTorch 1.8.1 on an NVIDIA GeForce RTX 2080Ti GPU. The segmentation performance was quantitatively measured by Dice score and Average Symmetric Surface Distance (ASSD) in 3D space. 

\begin{figure*}
\centering
\centerline{\includegraphics[width=17cm]{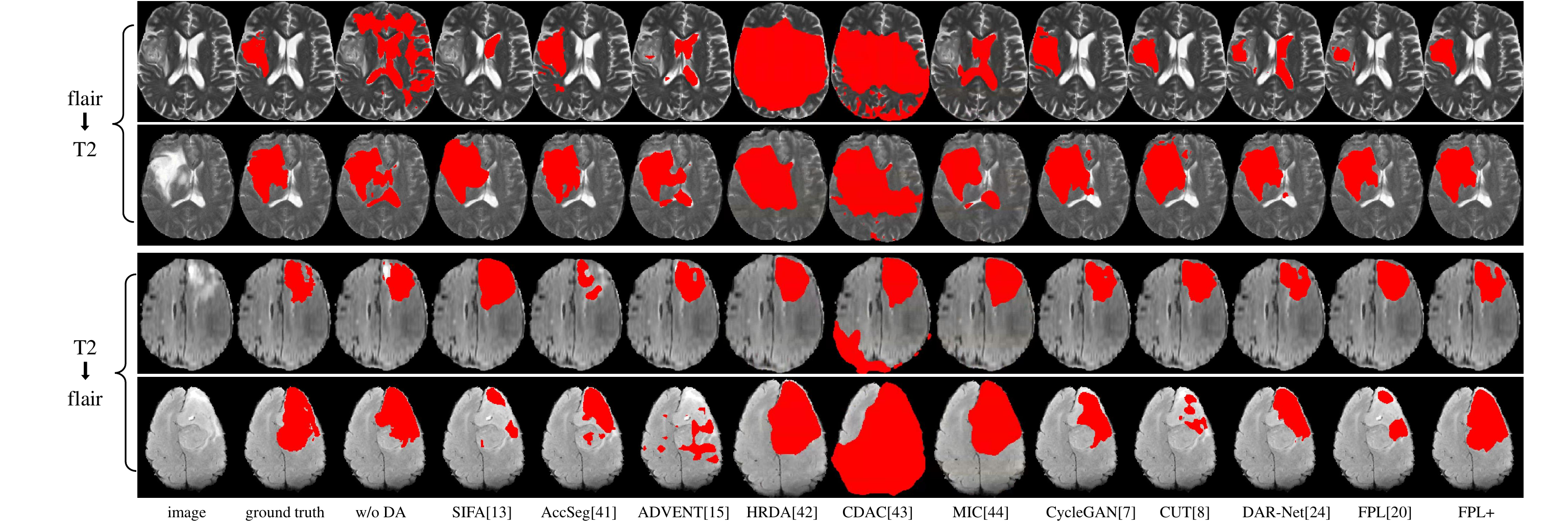}}
\caption{Visualization of segmentation results obtained by different UDA methods on the BraTs dataset. }
\label{fig:seg_bst}
\end{figure*}

\subsection{Comparison with SOTA Methods}
Our FPL+ was compared with ten state-of-the-art UDA methods:
 1) \textbf{CycleGAN}~\cite{CycleGAN2017} that performs unpaired image-to-image translation using cycle consistency loss and adversarial learning;
 2) \textbf{CUT}~\cite{park2020cut}  that maximizes mutual information between corresponding patches  using contrastive learning for image translation. We used CycleGAN and CUT to translate the labeled source-domain images into pseudo target-domain images to train a segmentor for the target domain, respectively.
 3) \textbf{SIFA}~\cite{chen2020unsupervised} that uses  synergistic image and feature alignment based on adversarial learning and a deeply supervised mechanism;
 4) \textbf{AccSeg}~\cite{zhou2021anatomy} that utilizes patch contrastive learning to adapt a segmentation network to a target imaging modality;
 5) \textbf{ADVENT}~\cite{vu2019advent} that combines entropy loss and adversarial loss for UDA in semantic segmentation.
6) 
 \textbf{HRDA}~\cite{hoyer2022hrda} that combines high-resolution and low-resolution crops to capture long-range context dependencies.
 7) \textbf{CDAC}~\cite{wang2023cdac} that proposes adaptation on attention maps with cross-domain attention layers.
 8) \textbf{MIC}~\cite{hoyer2023mic} that learns spatial context relations of the target domain as additional clues for robust visual recognition. \textcolor{black}{9}) \textbf{DAR-Net}~\cite{yao2022novel} that uses disentangled GAN for  image translation and employs a 3D CNN for segmentation. 
 \textcolor{black}{10}) \textbf{FPL}~\cite{wu2022fpl} that is a preliminary version of our FPL+, and it does not use dual-domain generator/segmentor and pixel-level weighting of pseudo labels.
 We also compared these methods with the ``\textbf{w/o DA}" lower bound, i.e., directly applying a model trained with $\mathcal{D}_s$ to images in $\mathcal{D}_t$, and with the ``\textbf{labeled target}", i.e., training the segmentation model in the target domain with full annotations. \textcolor{black}{They were also compared with  ``\textbf{strong upbound}" that means using labeled source-domain and target-domain images for training our dual-domain segmentation network. This serves as a theoretical upper bound for using the dual-domain data with full annotations for training.} Note that CycleGAN, CUT, DAR-Net, FPL and our FPL+ use 3D segmentation networks based on the same backbone of 2.5D CNN~\cite{wang2019automatic}, and the others use 2D segmentation networks that are coupled with their image/feature alignment process.

\subsubsection{Result of Vestibular Schwannoma Segmentation} 
We first performed bidirectional UDA between ceT1 and hrT2 on the VS dataset. Table~\ref{tab:sota_vs} shows the quantitative comparison of different methods in terms of Dice and ASSD. ``ceT1 to hrT2" means using ceT1 as the source domain and hrT2 as the target domain, respectively, while ``hrT2 to ceT1" is the opposite. 
The ``w/o DA" method obtained an average Dice of 0.00\% and 2.65\% in ``ceT1 to hrT2" and ``hrT2 to ceT1", respectively, indicating a significant domain shift between the two modalities. All the UDA methods showed improvements compared to w/o DA. SIFA~\cite{chen2020unsupervised} achieved an average Dice of 69.75\% and 67.48\% in the two directions, respectively. AccSeg~\cite{zhou2021anatomy} only achieved 30.92\% and 37.01\% respectively. 
\textcolor{black}{ADVENT~\cite{vu2019advent}, HRDA~\cite{hoyer2022hrda} and CDAC~\cite{wang2023cdac} only obtained slight performance improvement over "w/o DA". Despite that MIC~\cite{hoyer2023mic} was much better than these three methods on ``ceT1 to hrT2", it performed badly on ``hrT2 to ceT1", showing its low robustness in different cross modality settings.}

FPL achieved the highest performance among the existing methods, with Dice scores of 78.78\% and 84.90\% in the two directions, respectively.
The average Dice of our FPL+ was 82.92\% and 91.98\% in the two directions, respectively, and they were significantly higher than those of the other methods. \textcolor{black}{Note that for ``hrT2 to ceT1", our method was inferior to ``strong upbound",  but was even slightly better than ``labeled target” (91.98\% vs 90.72\% in terms of Dice), which was mainly due to that our segmentor leverages images from both domains for learning.} Fig.~\ref{fig:seg_vs} presents visual segmentation results of different methods. It shows that the other methods exhibit varying degrees of mis-segmentation, while our method closely aligns with the ground truth.

\begin{table}
  \centering
  \caption{Quantitative comparison of different UDA methods for bidirectional adaptation on glioma segmentation. $\dag$ indicates a significant improvement ($p$-value < 0.05) from the best values obtained by existing methods. }
  \resizebox{9cm}{!}{
    \begin{tabular}{c|c|c|c|c}
    \hline
    \multirow{2}[4]{*}{Method} & \multicolumn{2}{c|}{FLAIR to T2} & \multicolumn{2}{c}{T2 to FLAIR} \bigstrut\\
\cline{2-5}          & Dice (\%) & ASSD (mm) & Dice (\%) & ASSD (mm) \bigstrut\\
    \hline
    w/o DA & 47.16±24.39 & 20.82±11.31 & 68.46±21.74 & 8.71±8.38 \bigstrut[t]\\
    labeled target & 81.18±16.62 & 3.95±8.28 & 84.50±15.41 & 3.73±6.48 \\
    \textcolor{black}{strong upbound} & \textcolor{black}{81.26±16.91} & \textcolor{black}{3.84±6.61} & \textcolor{black}{86.69±11.96} & \textcolor{black}{2.30±2.01} \bigstrut[b]\\
    \hline
    SIFA~\cite{chen2020unsupervised}  & 55.52±20.30 & 14.77±9.06 & 66.03±14.34 & 7.45±4.38 \bigstrut[t]\\
    AccSeg~\cite{zhou2021anatomy} & 63.95±15.93 & 17.52±8.69 & 69.81±22.06 & 8.98±6.91 \\
    ADVENT~\cite{vu2019advent} & 39.83±24.07 & 16.76±8.43 & 55.03±23.34 & 10.51±8.79 \\
\textcolor{black}{HRDA}~\cite{hoyer2022hrda} & \textcolor{black}{27.48±18.39} & \textcolor{black}{27.52±10.31} & \textcolor{black}{63.06±14.65} & \textcolor{black}{13.63±6.37} \\
\textcolor{black}{CDAC}~\cite{wang2023cdac} & \textcolor{black}{25.55±14.11} & \textcolor{black}{33.61±10.24} & \textcolor{black}{21.40±9.83} & \textcolor{black}{38.96±7.88} \\
\textcolor{black}{MIC}~\cite{hoyer2023mic} & \textcolor{black}{49.23±28.12} & \textcolor{black}{12.48±9.27} & \textcolor{black}{76.23±8.44} & \textcolor{black}{3.83±1.36} \\
    CycleGAN~\cite{CycleGAN2017} & 66.66±24.20 & 7.26±8.20 & 75.47±23.86 & 4.57±8.05 \\
    CUT~\cite{park2020cut}   & 66.03±25.81 & 9.79±13.95 & 72.33±21.94 & 7.21±12.43 \\
    DAR-NET~\cite{yao2022novel} & 68.84±26.90 & 7.69±10.07 & 70.60±24.05 & 5.32±10.38 \\
    FPL~\cite{wu2022fpl} & 70.63±24.80 & 7.10±11.61 & 79.62±12.38 & 4.01±3.51 \\
    FPL+ (Ours) & \textbf{75.76±22.96\dag} & \textbf{4.46±5.74\dag} & \textbf{84.81±11.76\dag} & \textbf{2.72±2.70\dag} \bigstrut[b]\\
    \hline
    \end{tabular}%
    }
  \label{tab:sota_brats}%
\end{table}%

\begin{table*}[htbp]
  \centering
  \caption{\textcolor{black}{Quantitative comparison of different UDA methods for cardiac substructure segmentation. MRI and CT are used as the source and target domains,  respectively. $\dag$ indicates a significant improvement ($p$-value < 0.05) from the best values obtained by existing methods.}}
  \textcolor{black}{
  \resizebox{18cm}{!}{
    \begin{tabular}{c|c|c|c|c|c|c|c|c|c|c}
    \hline
    \multirow{2}[4]{*}{Method} & \multicolumn{4}{c}{Dice (\%)} &       & \multicolumn{4}{c}{ASSD (mm)} &  \bigstrut\\
\cline{2-11}          & AA    & LAC   & LVC   & MYO   & Average & AA    & LAC   & LVC   & MYO   & Average \bigstrut\\
    \hline
    w/o DA & 15.20±26.32 & 53.16±5.65 & 5.96±6.02 & 6.05±5.42 & 20.09±6.12 & 19.97±11.22 & 10.35±1.09 & 18.86±12.54 & 15.48±3.62 & 16.16±4.11 \bigstrut[t]\\
    labeled target & 95.49±1.99 & 85.86±12.38 & 78.34±25.47 & 79.67±17.98 & 84.84±14.10 & 0.57±0.37 & 1.21±0.95 & 1.50±1.42 & 1.28±0.94 & 1.14±0.91 \\
    strong upbound & 95.02±1.16     & 90.56±2.57     & 84.60±11.81     & 83.81±7.90     & 88.50±5.42     & 0.41±0.08     & 1.01±0.37     & 1.32±0.77     & 1.28±0.86     & 1.01±0.49 \bigstrut[b]\\
    \hline
    SIFA~\cite{chen2020unsupervised}  & 76.41±5.23 & 76.38±7.95 & 68.02±15.56 & 51.79±4.03 & 68.15±6.39 & 3.94±2.52 & 2.14±0.65 & 2.81±0.65 & 2.95±0.76 & 2.96±1.10 \bigstrut[t]\\
    AccSeg~\cite{zhou2021anatomy} & 58.96±9.31 & 72.46±2.73 & 67.21±8.08 & 59.21±4.34 & 64.46±4.89 & 7.37±3.72 & 4.47±0.82 & 6.45±1.75 & 5.42±1.28 & 5.93±1.27 \\
    ADVENT~\cite{vu2019advent} & 72.55±7.72 & 54.11±12.82 & 49.03±26.35 & 47.07±8.82 & 55.69±8.51 & 9.42±542 & 8.02±1.12 & 8.95±2.63 & 6.96±2.36 & 8.34±2.59 \\
\textcolor{black}{HRDA}~\cite{hoyer2022hrda} & \textcolor{black}{67.10±4.69} & \textcolor{black}{80.03±2.03} & \textcolor{black}{60.30±10.04} & \textcolor{black}{60.56±8.46} & \textcolor{black}{67.00±3.11} & \textcolor{black}{9.10±2.86} & \textcolor{black}{3.09±1.37} & \textcolor{black}{8.94±2.04} & \textcolor{black}{6.21±2.04} & \textcolor{black}{6.83±1.29} \\
\textcolor{black}{CDAC}~\cite{wang2023cdac} & \textcolor{black}{57.72±3.09} & \textcolor{black}{74.32±1.17} & \textcolor{black}{52.64±10.49} & \textcolor{black}{57.99±6.99} & \textcolor{black}{60.67±1.88} & \textcolor{black}{8.28±2.08} & \textcolor{black}{3.28±1.07} & \textcolor{black}{13.69±1.45} & \textcolor{black}{8.91±2.88} & \textcolor{black}{8.54±0.97} \\
\textcolor{black}{MIC}~\cite{hoyer2023mic} & \textcolor{black}{39.15±13.65} & \textcolor{black}{70.48±2.92} & \textcolor{black}{49.70±8.95} & \textcolor{black}{46.55±3.61} & \textcolor{black}{51.47±4.64} & \textcolor{black}{13.08±6.52} & \textcolor{black}{3.36±0.87} & \textcolor{black}{8.61±1.49} & \textcolor{black}{5.08±0.24} & \textcolor{black}{7.53±2.20} \\
    CycleGAN~\cite{CycleGAN2017} & 66.95±5.56 & 67.87±14.25 & 63.79±11.94 & 40.85±10.36 & 59.86±10.18 & 8.02±2.10 & 2.60±0.84 & 3.59±0.76 & 3.86±1.58 & 4.52±1.20 \\
    CUT~\cite{park2020cut}   & 41.06±24.46 & 76.94±3.31 & 74.85±7.49 & 54.08±14.94 & 61.73±5.01 & 4.09±1.60 & 2.83±0.72 & 3.52±0.55 & 3.36±0.86 & 3.45±0.34 \\
    DAR-NET~\cite{yao2022novel} & 70.11±7.36 & \textbf{82.24±3.16} & 59.28±34.27 & 59.34±17.41 & 67.74±1522 & 7.40±2.08 & 2.81±1.28 & 9.77±12.56 & 3.64±3.14 & 5.90±4.48 \\
    FPL~\cite{wu2022fpl} & \textbf{77.54±1.74} & 65.96±24.23 & 63.19±25.67 & 54.14±15.80 & 65.21±16.61 & \textbf{3.83±1.71} & 2.59±1.60 & 2.96±1.20 & 3.06±1.49 & 3.11±1.44 \\
    FPL+ (Ours) & 73.84±4.07 & 80.19±5.64 & \textbf{76.24±5.86} & \textbf{64.54±5.84} & \textbf{73.70±4.74\dag} & 3.90±1.94 & \textbf{1.89±0.54} & \textbf{2.34±0.28} & \textbf{2.33±0.83} & \textbf{2.61±0.86\dag} \bigstrut[b]\\
    \hline
    \end{tabular}%
    }
    }
  \label{tab:sota_mmwhs}%
\end{table*}%

\subsubsection{Results of Glioma Segmentation} 
Quantitative evaluation results of different UDA methods on the BraTS dataset are shown in Table~\ref{tab:sota_brats}. For ``FLAIR to T2", the Dice scores of ``w/o DA" were 47.16\%, indicating a certain domain gap between the two modalities. 
\textcolor{black}{The average Dice scores achieved by SIFA, AccSeg, and MIC were 55.52\%, 63.95\%, and 49.23\%, respectively. ADVENT, HRDA, and CDAC exhibited lower performance compared to "w/o DA", with scores of 39.83\%, 27.48\%, and 25.55\%, respectively.} Compared with these methods for 2D UDA, the other methods using 3D segmentation models obtained a better performance. Especially, FPL obtained the highest performance among the existing UDA methods with an average Dice of 70.63\% and ASSD of 7.10~mm. Our FPL+ further outperformed FPL~\cite{wu2022fpl}, with an average Dice and ASSD of 75.76\% and 4.46~mm, respectively.
In the ``T2 to FLAIR" direction,
the ``w/o DA" baseline obatained an average Dice of 68.46\%, and FPL obtained an average Dice of 79.62\%, which outperformed the other existing UDA methods.
\textcolor{black}{FPL+ achieved an average Dice of 84.81\% with an average ASSD of 2.72~mm, surpassing ``labeled target" and falling slightly behind the ``strong upper bound" that achieved a Dice of 86.69\% and ASSD of 2.30~mm.} This can be attributed to the inherently better contrast of FLAIR images of whole tumor and the ability of our method to extract rich domain-invariant information. A visual comparison between these methods is shown in Fig.~\ref{fig:seg_bst}, which demonstrates the superiority of our method for cross-modality UDA in different directions.  

\subsubsection{Results of Heart Segmentation} 
\begin{figure*}
    \centering
    \centerline{\includegraphics[width=18.0cm]{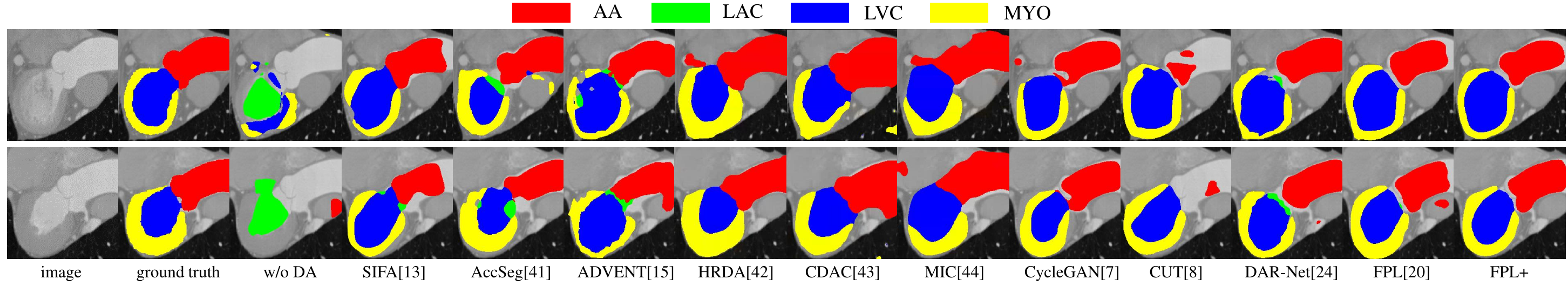}}
    \caption{\textcolor{black}{Visualization of segmentation results obtained by different UDA methods on the MMWHS dataset, where MRI and CT were used as source and target domains, respectively. }}
    \label{fig:seg_mmwhs}
\end{figure*}

\textcolor{black}{
Table~\ref{tab:sota_mmwhs} presents the quantitative evaluation results for various UDA methods on the heart segmentation dataset. The Dice scores of ``w/o DA" for different cardiac structures, including AA, LAC, LVC and MYO, were 15.20\%, 53.16\%, 5.96\%, and 6.05\%, respectively, and the average Dice score (20.09\%) was significantly lower than the ``labeled target" (84.84\%). This discrepancy indicates a notable domain gap between the MR and CT modalities.
\textcolor{black}{Comparatively, SIFA, AccSeg, ADVENT, HRDA, CDAC and MIC achieved average Dice scores of 68.15\%, 64.46\%, 55.69\%, 67.00\%, 60.67\%, and 51.47\%, respectively.} 
The average Dice scores for CycleGAN, CUT, and DAR-NET that utilize 3D segmentation models were 59.86\%, 61.73\%, and 67.74\%, respectively.
Given the anatomical consistency of the heart across different patients, FPL that only uses image-level weighting of pseudo labels and only leverages target-domain images for training obtained an average Dice score of 65.21\%. In contrast, our proposed method, FPL+ with additional pixel-level weighting to leverage dual-domain images for training, demonstrated superior performance with an average Dice and ASSD of 73.70\% and 2.61 mm, respectively, and it significantly outperformed the existing UDA methods. A visual comparison among these methods is depicted in Fig.~\ref{fig:seg_mmwhs}, demonstrating that our FPL+ achieves more accurate segmentation of heart sub-structures than the other methods.}

\begin{table*}[htbp]
  \centering
  \caption{Ablation study of our Dual Domain pseudo label Generator (DDG) on the VS dataset. Note that $\mathcal{D}_{s\rightarrow t}$ is a subset of $\mathcal{D}_{st}$, and $\mathcal{D}_{s}$ is a subset of $\mathcal{D}_{ss}$, respectively.  \textcolor{black}{$\dag$ means significant difference (p-VALUE < 0.05) from the final model in the Table. }
  }
    \begin{tabular}{ccccc|c|c|c|c}
    \hline
    \multirow{2}[4]{*}{$\mathcal{D}_{s\rightarrow t}$} & \multirow{2}[4]{*}{$\mathcal{D}_{s}$} & \multirow{2}[4]{*}{Dual-BN} & \multirow{2}[4]{*}{$\mathcal{D}_{st}$}  & \multirow{2}[4]{*}{$\mathcal{D}_{ss}$} & \multicolumn{2}{c|}{ceT1 to hrT2} & \multicolumn{2}{c}{hrT2 to ceT1} \bigstrut\\
    \cline{6-9}  &        &       &       &       & Dice (\%) & ASSD (mm) & Dice (\%) & ASSD (mm) \bigstrut\\
    \hline
    \checkmark            &       &       &       &       & 79.94±22.49\textcolor{black}{\dag} & 4.14±12.72\textcolor{black}{\dag} & 81.23±23.65\textcolor{black}{\dag} & 4.12±12.74\textcolor{black}{\dag} \bigstrut[t]\\
    \checkmark     & \checkmark            &       &       &       & 82.67±7.44\textcolor{black}{\dag} & 1.27±1.35 & 82.72±15.70\textcolor{black}{\dag} & 0.69±0.84 \\
    \checkmark     & \checkmark     & \checkmark     &       &        & 84.94±6.05 & 1.39±1.40 & 83.08±22.07 & 1.63±3.25\textcolor{black}{\dag} \\
     & \checkmark     & \checkmark         &  \checkmark     &       & 85.73±4.70 & \textbf{0.93±1.11} & 84.75±11.45 & \textbf{0.51±0.45} \\
      &      & \checkmark    &  \checkmark     & \checkmark     & \textbf{86.77±4.88} & 1.02±1.41 & \textbf{85.49±17.13} & 0.55±0.76 \bigstrut[b]\\
    \hline
    \end{tabular}%
  \label{tab:ablation_$DDG$}%
\end{table*}

\subsection{Ablation Study}
To validate each component in our FPL+, we conducted a comprehensive ablation study on the Dual-Domain pseudo label Generator (DDG) and the final segmentor $S$ using the VS  dataset.
For DDG, we investigated the effectiveness of our Dual-BN, $\mathcal{D}_{st}$ and $\mathcal{D}_{ss}$ based on CDDA. For training $S$, we investigated the effectiveness of Dual-BN, and image-level and pixel-level weighting. Our pseudo label filtering method was also compared with several existing noise-robust learning methods. It should be noted that all ablation study results were obtained from the validation set of the target domains.

\subsubsection{Effectiveness of CDDA and Dual-BN for DDG}
To investigate the effectiveness of CDDA and  Dual-BN, we first trained the pseudo label generator using $\mathcal{D}_{s \rightarrow t} = \{(X_i^{s \rightarrow t}, Y^s_i)\}$ as the baseline. As shown in Table~\ref{tab:ablation_$DDG$}, for ``ceT1 to hrT2", the baseline obtained an average Dice of 79.94\%. When using a combination of $\mathcal{D}_{s \rightarrow t}$ and $\mathcal{D}_{s}$ without dual-BN for training $G$, the average Dice was improved to 82.67\%, showing the benefit of combing images in the source and pseudo target domains for training. Introducing dual-BN further improved it to 84.94\%, demonstrating the effectiveness of using domain-specific batch normalization to deal with the domain shift for joint training. By introducing the auxiliary translator $T_{at}$, i.e., replacing $\mathcal{D}_{s\rightarrow t}$ by $\mathcal{D}_{st}$, the average Dice was 85.73\%. Finally, the proposed combination of $\mathcal{D}_{st}$, $\mathcal{D}_{st}$ and Dual-BN obtained the highest average Dice of 86.77\%, which shows superiority of the proposed CDDA. 

A similar conclusion can be obtained from the ``hrT2 to ceT1" direction, as shown in Table~\ref{tab:ablation_$DDG$}. The baseline of training from $\mathcal{D}_{s\rightarrow t}$ only obtained an average Dice of 81.23\%. Introducing $\mathcal{D}_s$ and dual-BN improved it to 82.72\% and 83.08\%, respectively. Using the dual-domain augmented images in $\mathcal{D}_{st}$ and $\mathcal{D}_{st}$ combined with dual-BN obtained an average  Dice score of 85.49\%, which outperformed the other variants.
\begin{table*}[htbp]
  \centering
  \caption{Comparison between different methods for training the final segmentor ($S$) on the VS segmentation dataset. The baseline is standard supervised learning from pseudo labels of target-domain images obtained by DDG. \textcolor{black}{$\dag$  means significant improvement ($p$-value < 0.05)  from the best values obtained by the three state-of-the-art methods}.}
    \begin{tabular}{cccccc|c|c|c|c}
    \hline
    \multirow{2}[4]{*}{Baseline} & \multirow{2}[4]{*}{$\mathcal{D}_{s}$} & \multirow{2}[4]{*}{Dual-BN} & \multirow{2}[4]{*}{Init from $G$} & \multirow{2}[4]{*}{$w$} & \multirow{2}[4]{*}{$M$} & \multicolumn{2}{c|}{ceT1 to hrT2} & \multicolumn{2}{c}{hrT2 to ceT1} \bigstrut\\
\cline{7-10}       &    &       &       &       &       & Dice (\%) & ASSD (mm) & Dice (\%) & ASSD (mm) \bigstrut\\
    \hline
         \checkmark  &   &    &       &       &       & 82.67±9.22 & 0.81±0.4 & 81.01±23.04 & 0.91±1.51 \bigstrut[t]\\
    \multicolumn{1}{c}\checkmark &\checkmark &       &       &       &       &   83.77±7.00 & 0.71±0.76 & 81.24±21.33 & 0.85±1.23 \\
    \multicolumn{1}{c}\checkmark &\checkmark & \checkmark     &       &       &  &      85.54±5.31 & 0.69±0.88 & 81.62±23.33 & 1.07±2.35 \\
    \multicolumn{1}{c}\checkmark &\checkmark & \checkmark     & \checkmark     &  &     &        87.21±4.56 & 1.06±1.42 & 83.32±22.03 & 0.65±1.22 \\
    \multicolumn{1}{c}\checkmark &\checkmark & \checkmark     & \checkmark     & \checkmark  &   &        88.01±4.03 & 0.36±0.08 & 85.63±21.65 & 0.66±1.66 \\
    \multicolumn{1}{c}\checkmark & \checkmark & \checkmark     & \checkmark     &       & \checkmark  &    88.15±4.50 & \textbf{0.34±0.08} & 85.15±17.05 & 0.50±0.85 \\
    \multicolumn{1}{c}\checkmark & \checkmark & \checkmark     & \checkmark     & \checkmark     & \checkmark    & \textbf{88.29±4.39}$\textcolor{black}{\dag}$ & 0.34±0.09$\textcolor{black}{\dag}$ & \textbf{86.57±14.41}$\textcolor{black}{\dag}$ & \textbf{0.46±0.75}$\textcolor{black}{\dag}$ \bigstrut[b]\\
    \hline
          
          & \multicolumn{5}{c|}{Co-teaching~\cite{han2018co}} & 83.93±7.69 & 2.08±2.56 & 81.19±20.87 & 2.60±2.88 \\
          & \multicolumn{5}{c|}{GCE Loss~\cite{zhang2018generalized}} & 84.14±6.48 & 0.83±0.50 & 83.78±14.15 & 1.87±2.51 \\
          & \multicolumn{5}{c|}{TriNet~\cite{zhang2020robust}}   & 85.86±3.69 & 1.12±2.11 & 84.18±15.37 & 1.39±1.79 \\
    \hline
    \end{tabular}%
  \label{tab:ablation_S}%
\end{table*}%

\subsubsection{Ablation Study for Training the Final Segmentor}
For ablation study of the final segmentor $S$, we set the baseline as standard supervised learning from pseudo labels of $D_t$ obtained by DDG, and gradually introduce the following components: 1) Adding the labeled images in $\mathcal{D}_s$ to the training set of $S$; 2) Using dual-BN for $S$ when jointly learning from $\mathcal{D}_t$ and $\mathcal{D}_s$; 3) Initializing $S$ from the trained $G$; 4) using the proposed image-level weighting based on size-aware uncertainty estimation; and 5) using the proposed pixel-level weighting based on dual-domain consensus.


As shown in Table~\ref{tab:ablation_S}, the Dice scores of the baseline for ``ceT1 to hrT2" and ``hrT2 to ceT1" were 82.67\% and 81.01\%, respectively. 
By additionally training with $\mathcal{D}_s$ and using dual-BN, the corresponding average Dice was increased to 85.54\% and 81.62\%, respectively. 
After applying initialization from $G$, the corresponding  Dice scores were further improved to 87.21\% and 83.32\%, respectively. 

When the image-level weight was used, the average Dice score was increased to 88.01\% for  ``ceT1 to hrT2" and 87.57\% for ``hrT2 to ceT1", respectively. It demonstrates that our image-level weight is useful in suppressing low-quality pseudo labels for robust learning.
Finally, when the proposed image-level weight and pixel-level weight map are combined to train the final segmentor, the resulting average Dice was 88.29\% for ``ceT1 to hrT2" and 86.57\% for ``hrT2 to ceT1", which outperformed the other variants. These results demonstrate that each component in our proposed method for training the final segmentor was effective.

\subsubsection{Comparison with Other Pseudo Label Learning Methods}
With the same set of pseudo labels generated by DDG for the target-domain training images $\mathcal{D}_t$, our proposed strategy to train $S$ was also compared with three state-of-the-art methods for learning from noisy labels:
1) \textbf{Co-teaching}~\cite{han2018co} that involves training two neural networks simultaneously, where each network selects high-quality pseudo labels based on the training loss within a mini-batch for the other; 2) \textbf{GCE Loss}~\cite{zhang2018generalized} that is a generalization of Mean Absolute Error (MAE) and cross entropy loss for robust learning; 3) \textbf{TriNet}~\cite{zhang2020robust} that employs three networks to iteratively select informative samples for training based on the consensus and discrepancy between their predictions.

The results of these methods are shown in the last three rows of Table~\ref{tab:ablation_S}. Co-teaching~\cite{han2018co} achieved  an Dice score of 83.93\% for ``ceT1 to hrT2" and 81.19\% for ``hrT2 to ceT1", respectively. The GCE loss~\cite{zhang2018generalized} obtained a higher average Dice score of 84.14\% for ``ceT1 to hrT2" and 83.78\% for ``hrT2 to ceT1", respectively. The corresponding value obtained by TriNet~\cite{zhang2020robust} was 85.86\% for ``ceT1 to hrT2" and 84.18\% for ``hrT2 to ceT1", respectively. Note that the performance of these methods was lower than that of ours.

\subsubsection{Effectiveness of Hyper-parameters}
\begin{figure}
    \centering
    \centerline{\includegraphics[width=8cm]{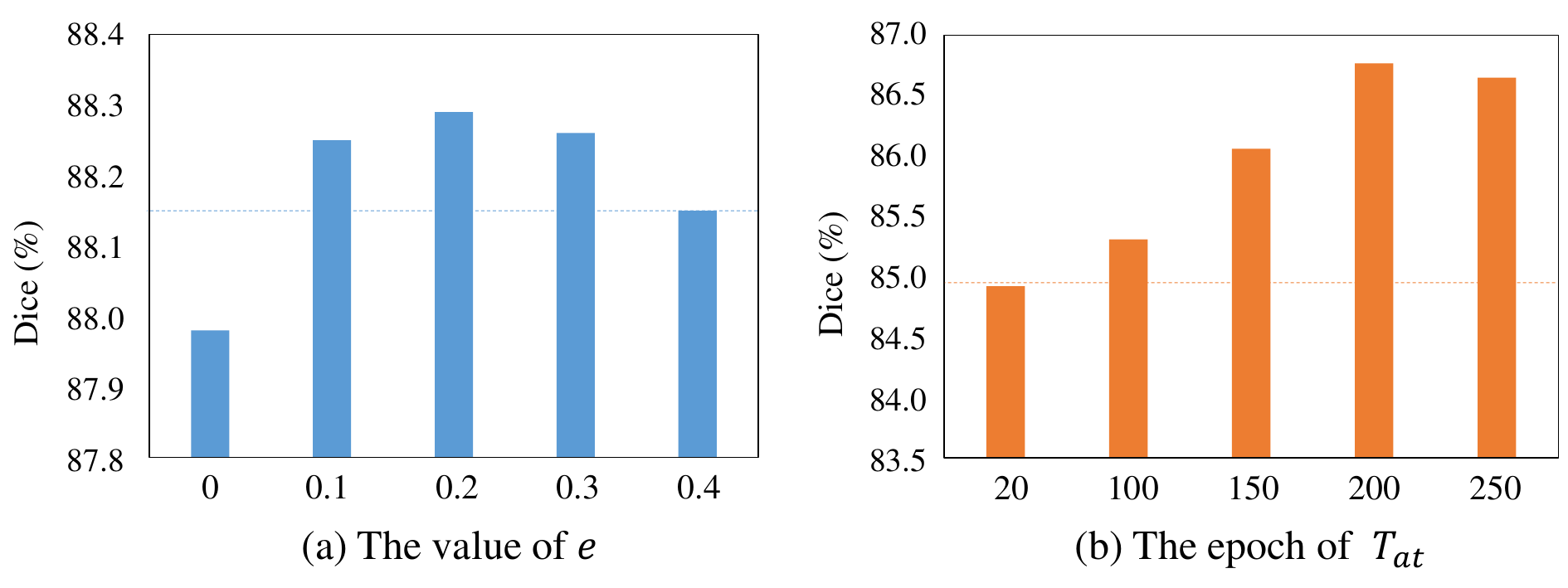}}
    \caption{Sensitivity analysis with respect to hyper-parameters $e$ and epoch for auxiliary target style translator. The dashed lines represent the results obtained without using image-level weighting in (a) and without employing an auxiliary style translator in (b), respectively.}
    \label{fig:value_e_epoch}
\end{figure}

Our method has two core hyper-parameters: threshold $e$ on the entropy map for image-level weighting and epoch number for selecting the auxiliary translator. To explore the impact of $e$, we varied its value from 0 to 0.4. The performance on the validation set of hrT2 on VS dataset is shown in Fig.~\ref{fig:value_e_epoch}. 
It's clear that with $e = 0$, which means using all pixels in the volume to normalize $v_{j}$, the performance is inferior to that with other $e$ values. 
The performance improved when $e$ was set from 0.1 to 0.3, and we can find that the performance was relatively stable when $e$ changes from 0.1 to 0.3, with the highest results achieved at $e = 0.2$. Therefore, we set $e = 0.2$ for our method. 

Then, the influence of the training epochs for the auxiliary target style translator $T_{at}$ on the quality of pseudo-label generation was investigated. Fig.~\ref{fig:value_e_epoch} shows results on the hrT2 validation set. Setting the epoch number of $T_{at}$ to  20 led to a  Dice of 84.92\%, which was slightly lower than not using the auxiliary translator (84.94\%), indicating that $T_{at}$ with a small epoch number does not help to improve the pseudo label generator. In contrast, when the epoch number increased to 100 and 150, the Dice obtained by pseudo label generator was improved to 85.31\% and 86.06\%, respectively. At epoch 200, the performance reached its peak at 86.77\%. However, at epoch 250, as the $T_{at}$ became similar to the primary translator, the diversity of augmented images would be reduced, and the corresponding Dice was slightly reduced to 86.65\%.

\begin{figure}
    \centering
    \centerline{\includegraphics[width=7.5cm]{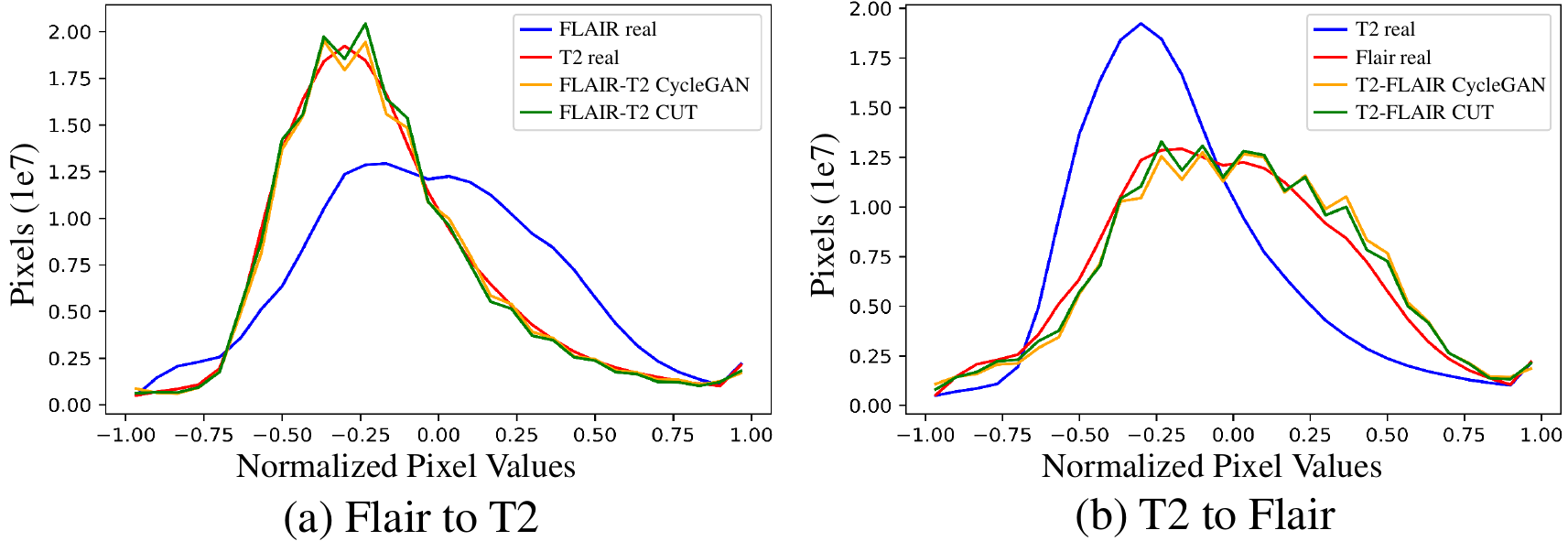}}
    \caption{{Visual comparison of intensity distribution of FLAIR and T2 domain images in the BraTS dataset, alongside pseudo-target images translated by CycleGAN and CUT. }}
    \label{fig:his_map}
\end{figure}

\section{Discussion}\label{sec:discussion}
For cross-modality unsupervised adaptation, image alignment to reduce the domain gap and selecting reliable pseudo labels are two key factors for obtaining good performance of our method. First, our cross-domain data augmentation using CycleGAN can effectively align the two domains at the image level. Fig.~\ref{fig:his_map} compares the intensity distribution of source domain, target domain and the augmented source and target domains. It can be observed that when converted from FLAIR to T2-weighted imaging, the augmented images are well aligned with real T2 images, and the same conclusion can be obtained when converting T2-weighted images to FLAIR. This does not only makes the two domains well aligned, but also provides more labeled samples for both domains, leading to higher performance of the segmentation model on the target domain. 
Second, the image-level and pixel-level weighting selects reliable pseudo labels for training the final segmentor. For tumors with various sizes and shapes, the pseudo label quality varies largely on different samples in the target domain, and image-level weighting can effectively reject low-quality pseudo labels, especially for cases with small irregular shapes and appearances. However, for organs with limited variations of shape and appearance in the target domain (eg., the heart), the pseudo labels have similar quality at the image level, our image-level weighting struggles to demonstrate a pronounced advantage, and the pixel-level weighting demonstrates more effectiveness in dealing with such scenarios. 

Additionally, our method focuses on 3D medical images, and due to GPU memory constraints and artifacts introduced by patch-wise image translation, achieving end-to-end image generation and segmentation on 3D images is challenging. As a result, our method involves two steps that are for training the pseudo label generator and the final segmentor respectively, which adds a certain level of complexity. However, as the final segmentor is initialized from the pseudo label generator, the former requires fewer training epochs.
Furthermore, our method requires that the segmentation targets should be visible with similar topologies in the source and target domains. For instance, we experimented with UDA between FLAIR and T2 images as both of them can show the whole tumor region, but applying our method to UDA between FLAIR and ceT1 may not be suitable, as ceT1 is less effective to visualize the peritumoral edema region. 

For model complexity, our method has two translators (11.366$M$ for each), two discriminators (2.763$M$ for each), and the model size of both pseudo label generator $G$ and final segmentator $S$ is 30.708$M$. Note that $S$ is initialized by $G$. Compared with using  CycleGAN for image translation followed by a segmentor in the target domain, our method only introduces extra BN layers, leading to a slight increase of model size of 0.012$M$. Due to the cross-domain data augmentation, our method has more augmented images for training $G$ and $S$, and they take 32.7 hours and 6.5 hours on the VS dataset, respectively, compared with  13.2 hours for the segmentor used after CycleGAN. Despite this, both methods only use the segmentor for inference, and share an identical inference time of 0.43 seconds per 3D volume, which is efficient for testing.

\section{Conclusion}\label{sec:conclusion}
In this paper, we propose an enhanced version of the Filtered Pseudo Label (FPL)-based cross-modality unsupervised domain adaptation method, called FPL+, for 3D medical image segmentation.
To generate high-quality pseudo labels in the target domain, we first propose a Cross-Domain Data Augmentation (CDDA) approach to augment the labeled source-domain images into a dual-domain training set consisting of a pseudo source-domain set and a pseudo target-domain set. The dual-domain augmented images are used to train a Dual-Domain pseudo label Generator (DDG), which incorporates domain-specific batch normalization layers to learn from the dual-domain images while dealing with the domain shift effectively.  
To enhance the performance of the final segmentor, we propose joint training from the labeled source-domain images and target-domain images with pseudo labels, and to deal with noisy pseudo labels, image-level weighting based on size-aware uncertainty and pixel-level weighting based on dual-domain consensus are proposed. 
The results on three public multi-modality datasets for brain tumor and whole heart segmentation show that our method outperforms existing UDA methods, and can even surpass fully supervised learning on the target domain in some cases. In the future, it is of interest to apply our method to other segmentation tasks.

\bibliographystyle{IEEEtran}
\bibliography{myref}
\end{document}